\documentclass[sigconf]{acmart}

\usepackage{tikz}
\usepackage{algorithm}
\usepackage{algpseudocode}
\usetikzlibrary{positioning,fit,backgrounds,arrows.meta}
\usepackage{xcolor}
\AtBeginDocument{%
  }

\begin{document}

\title{On Accelerating Grounded Code Development for Research}

\author{
Santosh Ganji \\
\texttt{sg@nirvawireless.com}
}


\renewcommand{\shortauthors}{Santosh Ganji}

\begin{abstract}

A major challenge for niche scientific and technical domains in leveraging coding agents is the lack of access to up-to-date, domain-specific knowledge. Foundational models often demonstrate limited reasoning capabilities in specialized fields and cannot inherently incorporate knowledge that evolves through ongoing research and experimentation. Materials scientists exploring novel compounds, communication engineers designing and evaluating new protocols, and bioengineering researchers conducting iterative experiments all face this limitation. These experts typically lack the resources to fine-tune large models or continuously embed new findings, creating a barrier to adopting AI-driven coding agents. To address this, we introduce a framework that gives coding agents instantaneous access to research repositories and technical documentation, enabling real-time, context-aware operation. Our open-source implementation allows users to upload documents via \href{https://doc-search.dev}{doc-search.dev} and includes \href{https://shotsan.github.io/zed-custom/}{zed-fork}, which enforces domain-specific rules and workflows. Together, these tools accelerate the integration of coding agents into specialized scientific and technical workflows.

\end{abstract}
\maketitle


\section{Introduction}

Large language models (LLMs) \cite{openai2023gpt4} have demonstrated remarkable capabilities in natural language understanding, code generation, and reasoning. However, leveraging these models in specialized scientific and technical domains remains a significant challenge. Domain experts—such as materials scientists, communication engineers, and bioengineers—frequently work with rapidly evolving knowledge, including experimental results, novel protocols, and technical documentation. Foundational LLMs are limited in their ability to reason over such dynamic, domain-specific information, and fine-tuning large models to capture these evolving datasets is often impractical due to computational costs and required expertise. Consequently, researchers face barriers in integrating LLMs effectively into their workflows, limiting the potential impact of AI-driven tools in scientific research.

Prior approaches have sought to improve LLM reasoning and knowledge integration through structured reasoning, tool augmentation, and knowledge grounding. Agentic reasoning frameworks, such as ReAct \cite{yao2023react}, interleave reasoning steps with actions like querying external APIs, enabling iterative information retrieval. Tool-augmented models, exemplified by Toolformer \cite{schick2023toolformer}, allow LLMs to autonomously decide when and how to use external tools during generation. Knowledge-graph-based retrieval augmented generation \cite{lewis2020rag} (KG-RAG) systems combine structured knowledge graphs with retrieval pipelines to support multi-hop reasoning. While these methods demonstrate improved reasoning in controlled settings, they encounter limitations when applied to real-world scientific workflows.

Agentic reasoning frameworks can accumulate errors over long sequences of actions, leading to inconsistent outcomes. Tool-based approaches often fail when domain grounding is insufficient, and knowledge-graph systems require substantial manual construction and ongoing maintenance, which is impractical in rapidly evolving research environments. Additionally, all these methods tend to assume relatively static knowledge bases, whereas scientific workflows are highly dynamic and heterogeneous, involving experimental datasets, code repositories, and technical documentation that change continuously.

Lowering barriers to integrate documentation is critical for enabling practical adoption of coding and analysis agents. Researchers should be able to leverage LLMs to automate routine tasks, prototype experiments, and analyze data without extensive machine learning expertise or costly model retraining or complex RAG methods. By freeing domain experts from auxiliary implementation details, intelligent agents can allow them to focus on experimental design, hypothesis testing, and knowledge discovery, accelerating both research and innovation.

Despite the availability of advanced embedding-based retrieval methods, we argue that simple lexical search is sufficient for grounding coding agents in domain-specific artifacts at this stage. Embedding based approaches can capture semantic similarity across documents but often require significant preprocessing, indexing, and tuning, which can slow iteration and reduce responsiveness in fast-moving research environments. In contrast, lexical search provides immediate, deterministic access to relevant documents and datasets, ensuring that agents can retrieve up-to-date information quickly and reliably. By prioritizing speed and reproducibility, our framework allows scientists and engineers to rapidly integrate LLMs into their workflows while leaving open the possibility of incorporating more sophisticated retrieval strategies in future iterations.

While complex reasoning in specialized scientific domains may remain an open and unsolved challenge, it is crucial to first lower the barrier for using coding and analysis agents. Researchers should be able to leverage intelligent agents to automate routine coding tasks, perform rapid data analysis, and prototype experiments without requiring extensive machine learning expertise or costly model fine-tuning. By providing immediate access to relevant repositories, technical documentation, and experimental datasets, such agents can serve as effective collaborators, enabling scientists and engineers to focus on research design, hypothesis exploration, and interpretation of results, rather than on auxiliary implementation details. Removing these barriers accelerates experimentation and knowledge discovery, creating a foundation upon which more sophisticated reasoning capabilities can later be integrated.

To address these limitations, we propose a framework that enables coding agents to directly access dynamic research repositories and experimental artifacts, providing real-time contextual grounding without requiring large-scale model retraining or manual knowledge graph construction. We also introduce new tools and a mechanism to update the system prompts of the coding agents to leverage these tools and conduct research effectively. We also demonstrate a case study on wireless standards documentation and show  better leverage the

\section{Preliminaries}
\label{sec:preliminaries}

This section reviews three retrieval paradigms relevant to retrieval augmented systems: {Retrieval‑Augmented Generation (RAG)}, {Knowledge Graph‑based RAG (KG‑RAG)} and Lexical Search. For each method, we describe the mechanisms for embedding creation, retrieval, compute and scalability trade‑offs, and known failure modes in large domain‑specific corpora.

\subsection{Retrieval‑Augmented Generation (RAG)}

Retrieval‑Augmented Generation (RAG) is an architecture in which queries are answered by retrieving relevant external documents from a corpus and using them as context for downstream tasks. RAG systems decouple knowledge access from model parameters by employing a vector database of text chunks that can be queried efficiently at run time. In standard RAG pipelines, the core building blocks are text chunking, dense vector embedding, and vector similarity search. Classic RAG has been a foundation for many extensions and variants due to its ability to provide up‑to‑date information to downstream components.

A fundamental component of RAG is \textit{text chunking}. Documents are partitioned into smaller text segments that fit within the input limits of embedding models. For example, a typical scientific paper with 7,000–15,000 tokens might be split into 10–30 chunks of 500–1,000 tokens each. Each chunk is treated as an independent semantic unit for retrieval. The choice of chunk size affects both retrieval granularity and storage requirements: smaller chunks improve precision by isolating local context, whereas larger chunks reduce the total number of vectors but may blur fine details.

Once text is chunked, each segment is mapped to a dense vector in a high‑dimensional space using an embedding model. Modern embedding models such as DPR or Sentence Transformers produce vector representations where geometric similarity approximates semantic similarity. These vectors are stored in a vector database where similarity search can be performed efficiently at query time. Since embeddings are computed offline, embedding creation is expensive when the corpus is large: embedding millions of chunks with large models requires substantial GPU compute and must be repeated whenever the corpus changes.

\paragraph{Retrieval Methods} Vector similarity search can be implemented using exact or approximate methods. Exact nearest neighbor search has linear complexity $\mathcal{O}(N)$ with respect to the number of vectors $N$, making it infeasible for large corpora. To address this, RAG systems typically employ approximate nearest neighbor indexes such as Hierarchical Navigable Small World (HNSW) graphs or Inverted File (IVF) structures, which achieve sub‑linear query time in practice. These methods trade off retrieval accuracy for speed and memory efficiency.

\paragraph{Compute and Scalability Trade‑offs}  
Dense embedding creation is the primary compute bottleneck in RAG systems; each text chunk requires a forward pass through an embedding model. For repositories comprising tens of millions of chunks, embedding computation alone can require hundreds of GPU‑hours. Approximate search indexes such as HNSW reduce query latency but incur significant memory overhead because they maintain additional graph or clustering metadata proportional to $N$. Frequent updates to the document corpus necessitate periodic recomputation of embeddings and index rebuilding, which scales poorly with corpus size.

\paragraph{Failure Modes}  
Approximate retrieval can miss relevant chunks due to the inherent trade‑off between speed and recall. Domain specific terminology or subtle distinctions common in scientific text may not be well captured by general embedding models, resulting in low retrieval precision. Additionally, vector similarity search does not explicitly capture structural relationships between terms, leading to noisy retrieval results when semantic similarity alone is insufficient to distinguish domain‑specific relevance.

\begin{table*}[h!]
\centering
\caption{Comparison of Major Approximate Nearest Neighbor (ANN) Methods for RAG}
\label{tab:ann-methods-extended}
\begin{tabular}{p{2.cm} p{4cm} p{2cm} p{2cm} p{2cm} p{4cm}}
\toprule
\textbf{Method} & \textbf{Mechanism} & \textbf{Complexity} & \textbf{Memory} & \textbf{Dynamic Updates} & \textbf{Pros / Cons / Use Cases} \\
\midrule
HNSW (Graph-based) & Multi-layer proximity graph; nodes = vectors, edges = nearest neighbors; hierarchical search & Sub-linear in practice & High; stores embeddings + graph edges & Supports incremental insertions; may require periodic graph repair for high recall & Pros: High recall, fast; good for dynamic corpora. Cons: Memory-intensive; graph can degrade with many insertions. Use: Scientific literature, semantic QA systems \\
\midrule
IVF (Clustering-based) & Partition vector space into clusters; vectors assigned to nearest cluster; search probes subset of clusters & Sub-linear; tunable via `nprobe` & Moderate; stores embeddings + centroids/inverted lists & Limited incremental support; adding many vectors may require centroid retraining & Pros: Scales to very large corpora; memory-efficient. Cons: Approximate; cluster drift over time. Use: Large-scale document/patent retrieval \\
\midrule
LSH (Hashing-based) & Hash vectors into buckets; similar vectors collide; query examines matching buckets & Sub-linear; depends on hash function & Low to moderate; stores hash codes & Supports incremental inserts easily, hash function may need tuning & Pros: Very fast; simple to implement. Cons: Lower recall in high-D; sensitive to hash parameters. Use: Medium-scale datasets, fast filtering \\
\midrule
Tree-based (Annoy, kd-tree, Ball-tree) & Recursive space partitioning into trees; search traverses branches & Logarithmic to linear depending on tree depth & Low to moderate; stores tree nodes + vectors & Limited; adding many vectors may require rebuilding tree & Pros: Simple, memory-efficient. Cons: Poor performance in high-dimensional embeddings (>1k dims). Use: Medium-size corpora, mid-dimensional embeddings \\
\midrule
PQ / IVF-PQ (Quantization-based) & Compress vectors into low-bit codes within clusters; search uses quantized distances & Sub-linear; faster than full IVF & Low; stores compressed vectors + cluster metadata & Limited; adding new vectors may require re-quantization & Pros: Very memory-efficient; scales to billions of vectors. Cons: Approximate; quantization introduces errors. Use: Extremely large-scale retrieval (e.g., patent corpora, web-scale corpora) \\
\bottomrule
\end{tabular}
\end{table*}

\subsection{Knowledge Graph‑based RAG (KG‑RAG)}

Knowledge Graph‑based Retrieval‑Augmented Generation extends RAG by integrating structured entity and relation information alongside vector embeddings. In KG‑RAG systems, documents are first parsed to extract entities and relations, which form a knowledge graph representing semantic connections among domain concepts. Retrieval then exploits both vector similarity and structured relationships from the graph to select relevant information. Recent work has shown that combining structured graph retrieval with dense text retrieval improves precision in reasoning‑oriented tasks, especially in specialized domains that require multi‑hop relational understanding.

In practice, KG‑RAG pipelines begin with entity and relation extraction from text. This process adds preprocessing overhead and often requires dedicated models for entity recognition particular to the domain. Chunks of text are embedded as in standard RAG, and entities are linked to graph structures that encode semantic relationships (such as subject‑predicate‑object triples). Retrieval mechanisms then combine vector similarity with entity‑based filters or graph traversal methods to prioritize chunks that align not only semantically but also relationally with the query.

\paragraph{Retrieval Methods} KG‑RAG retrieval operates in a hybrid fashion. Initial candidate selection is performed via dense vector similarity, followed by reranking or filtering based on graph connectivity, entity co‑occurrence, or subgraph relevance scores. Some KG‑RAG approaches work by identifying multi‑hop paths in the graph that maximize relevance to a query. This hybrid retrieval strategy improves the ability to capture structured context and relationships that pure vector search might overlook.

\paragraph{Compute and Scalability Trade‑offs}  
KG‑RAG incurs additional compute costs beyond RAG due to entity extraction and knowledge graph construction. Entity and relation extraction require domain‑adapted NLP models and add preprocessing overhead. The hybrid index must maintain both vector and graph data structures, increasing memory usage. Frequent updates to the knowledge graph, such as the addition of new entities or relations, necessitate reindexing both the graph and associated vectors, further increasing compute demands.

\paragraph{Failure Modes}  
Entity extraction errors can propagate to retrieval, reducing precision. Graph structures can grow large, particularly in complex scientific domains, making traversal or subgraph scoring computationally expensive. Moreover, hybrid retrieval requires careful balancing between semantic similarity and structural relevance; overly rigid graph filters may exclude relevant but semantically close text, while overly permissive vector search reintroduces noise.

\subsection{Text Chunking and Vector Storage}

Scientific documents often exceed the token limits of embedding models, requiring chunking prior to embedding. Each chunk generates one dense vector of $d$ dimensions, where $d$ may range from several hundred to over a thousand. Vector storage scales linearly with the number of chunks. For example, splitting a corpus of 1 million scientific papers into 20 chunks each yields 20 million vectors, which at 1,536 dimensions and 4 bytes per dimension requires on the order of 120\,GB of storage. Frequent updates increase both embedding and index maintenance costs.

\subsection{Summary of Limitations}

Both RAG and KG‑RAG enable scalable access to external knowledge, but face significant limitations in large, dynamic, domain specific corpora. Approximate nearest neighbor search introduces trade‑offs between speed, memory, and recall. Embeddings trained on general corpora may fail to capture technical domain nuances. KG‑RAG improves relational retrieval but incurs additional preprocessing, indexing, and update overhead. These limitations motivate structured retrieval approaches that can operate efficiently on large scientific datasets.

\subsubsection{Approximate Nearest Neighbor Retrieval with FAISS}

In retrieval systems used for RAG, the core task is to find vectors in a large database that are similar to an embedded query vector. Exact nearest neighbor search computes the distance between the query and every stored vector, which has linear complexity \(\mathcal{O}(N)\) and becomes infeasible for large \(N\). To scale to millions or billions of vectors, FAISS ({Facebook AI Similarity Search}) \cite{johnson2019faiss} implements approximate nearest neighbor (ANN) methods that reduce search cost by constructing data structures enabling sub‑linear traversal of the vector space. Two widely used ANN techniques in FAISS are {Inverted File (IVF)} and \textbf{Hierarchical Navigable Small World (HNSW)} indexing.

\paragraph{Embedding Creation and Initial Indexing}  
Given a corpus of text chunks (e.g., segments from scientific papers), each chunk is first converted into a dense vector using an embedding model. For example, a 1,536‑dimensional embedding model produces a 6KB float vector per chunk. Once all vectors are generated, they are added to an index structure that supports efficient retrieval. FAISS supports multiple index types; building the index transforms the embedding set into a form that enables fast lookup during queries. The specific index construction differs between IVF and HNSW, as described below.

\paragraph{Inverted File (IVF) Indexing}  
IVF organizes vectors by partitioning the high‑dimensional space into a set of coarse clusters using a clustering algorithm such as k‑means. During the training phase, FAISS computes a set of \(n_{\text{list}}\) cluster centroids over the vector dataset. Each vector is then assigned to its nearest centroid, forming an inverted list of vectors for that cluster. At query time, the query vector is compared against the centroids, and only a subset of clusters (controlled by the hyperparameter \(n_{\text{probe}}\)) is selected for detailed scanning. This reduces the number of vectors examined from \(N\) to roughly \(N \times (n_{\text{probe}}/n_{\text{list}})\), yielding sub‑linear empirical retrieval cost. Choosing a high number of centroids and probing more lists increases recall but also increases compute per query. If the true nearest neighbor resides in a cluster not selected for probing, it will not be retrieved, which is a fundamental approximation trade‑off in IVF.

\paragraph{Hierarchical Navigable Small World (HNSW) Indexing}  
HNSW builds a multi‑layer proximity graph over the vectors. Each node in the graph corresponds to one vector, and edges connect it to a predefined number \(M\) of its neighbors. The graph is organized hierarchically: upper layers contain a sparse subset of nodes with long‑range connections, while lower layers contain all nodes with denser local connectivity. During retrieval, the search begins at an entry point in the highest layer and greedily traverses edges toward vectors closer to the query, gradually descending through the layers until approximating the nearest vectors in the bottom layer. A parameter `efSearch` controls the size of the candidate queue explored during this traversal: a larger `efSearch` increases recall but also increases latency. HNSW does not require clustering or training, but its graph connectivity and memory overhead scale with the number of vectors and the number of connections per node.

\paragraph{Example}  
Suppose we have indexed 10 million chunk vectors for a large scientific repository. In an IVF index with \(n_{\text{list}}=65{,}536\) clusters, each vector is associated with one centroid. A query vector might first compute its distance to all 65,536 centroids to find the top \(n_{\text{probe}}=32\) clusters, then only examine the vectors in those lists. In contrast, an HNSW index with connectivity \(M=32\) would represent each vector as a node with 32 neighbor links. During a query with `efSearch=64`, the algorithm traverses a small subset of graph nodes (often <1 \% of the dataset) to identify approximate nearest neighbors. Both methods achieve sub‑linear querying compared to a flat exhaustive scan, but they make different trade‑offs between speed, memory, and recall.

\paragraph{Trade‑offs and Parameter Tuning}  
IVF requires a clustering/training phase before vectors can be added, and its recall depends heavily on the choice of \(n_{\text{list}}\) and \(n_{\text{probe}}\). A small \(n_{\text{probe}}\) yields faster queries but risks missing relevant vectors outside the probed clusters. Increasing \(n_{\text{probe}}\) improves recall at the cost of scanning more vectors per query. HNSW, by contrast, does not require a training step but requires careful tuning of connectivity \(M\) and search breadth `efSearch`: higher values improve recall but raise memory consumption and query time. In both methods, indexing structures must be maintained or rebuilt when vectors are frequently added or removed, imposing scalability challenges for dynamic corpora.

\paragraph{Failure Modes}  
Approximate retrieval methods inherently trade recall for speed. In IVF, if clustering poorly aligns with the underlying vector geometry, relevant vectors may be assigned to distant clusters and never probed. In HNSW, insufficient graph connectivity or low `efSearch` can result in paths that fail to reach true nearest neighbors. Additionally, embeddings that do not well represent domain‑specific semantics (e.g., specialized scientific terminology) may reduce the effectiveness of nearest neighbor proximity measures, further degrading retrieval quality. These factors highlight the practical limitations of ANN methods when applied to large, evolving scientific datasets and motivate careful tuning and evaluation of index parameters.

\begin{table*}[h!]
\centering
\caption{Comparison of Lexical and Semantic Retrieval Methods}
\label{tab:retrieval-comparison}
\begin{tabular}{p{3cm} p{6cm} p{6cm}}
\toprule
\textbf{Aspect} & \textbf{Lexical Retrieval} & \textbf{Semantic Retrieval} \\
\midrule
Basis & Exact token match; words largely treated in isolation & Meaning-based similarity; embeddings capture context and relationships \\
\midrule
Models & BM25, TF-IDF & BERT, Sentence-BERT, SciBERT, GPT embeddings \\
\midrule
Pros & Fast, simple, scalable; works well for keyword-based search & Captures synonyms, paraphrases, and context; can handle semantic similarity across different phrasings \\
\midrule
Cons & Misses meaning beyond exact words; cannot handle synonyms or paraphrases; insensitive to word order or context & Computationally heavier; memory-intensive; requires ANN structures for large corpora; approximate retrieval may miss some results \\
\midrule
Example & Query \texttt{"car acceleration"} $\rightarrow$ retrieves only documents containing \texttt{"car"} and \texttt{"acceleration"} & Query \texttt{"car acceleration"} $\rightarrow$ retrieves \texttt{"automobile speed"} or \texttt{"vehicle acceleration"} even if exact words differ \\
\bottomrule
\end{tabular}
\end{table*}

\subsection{Lexical Search}

Lexical search is a traditional information retrieval (IR) approach that retrieves documents based on the presence and frequency of query terms in documents. Unlike semantic search methods that rely on vector embeddings or deep learning models, lexical search operates directly on the surface form of words.

In lexical search, a user query is decomposed into tokens and matched against tokens present in indexed documents. Documents are ranked based on statistical measures such as term frequency and inverse document frequency. One of the most widely used ranking functions in lexical search systems is the BM25 scoring function.

Let $q$ be a query and $D$ be a document. The BM25 score between them is defined as:

\[
\text{score}(D, q) = \sum_{t \in q} IDF(t) 
\cdot
\frac{f(t,D)\,(k_1 + 1)}
{f(t,D) + k_1 \left(1 - b + b \frac{|D|}{avgdl}\right)}
\]

where:

Lexical search is widely used due to its efficiency, interpretability, and strong performance for keyword-based queries.

\subsubsection{Building a Lexical Search System}

A lexical search system typically consists of several stages:

\paragraph{Document Collection}
A corpus of documents is gathered. These documents may be web pages, research articles, product descriptions, or other textual data.

\paragraph{Text Processing}
Before indexing, documents undergo preprocessing steps such as:

\begin{itemize}
\item Tokenization
\item Lowercasing
\item Stopword removal
\item Stemming or lemmatization
\end{itemize}

\begin{figure*}
    \centering
    \includegraphics[width=1\linewidth]{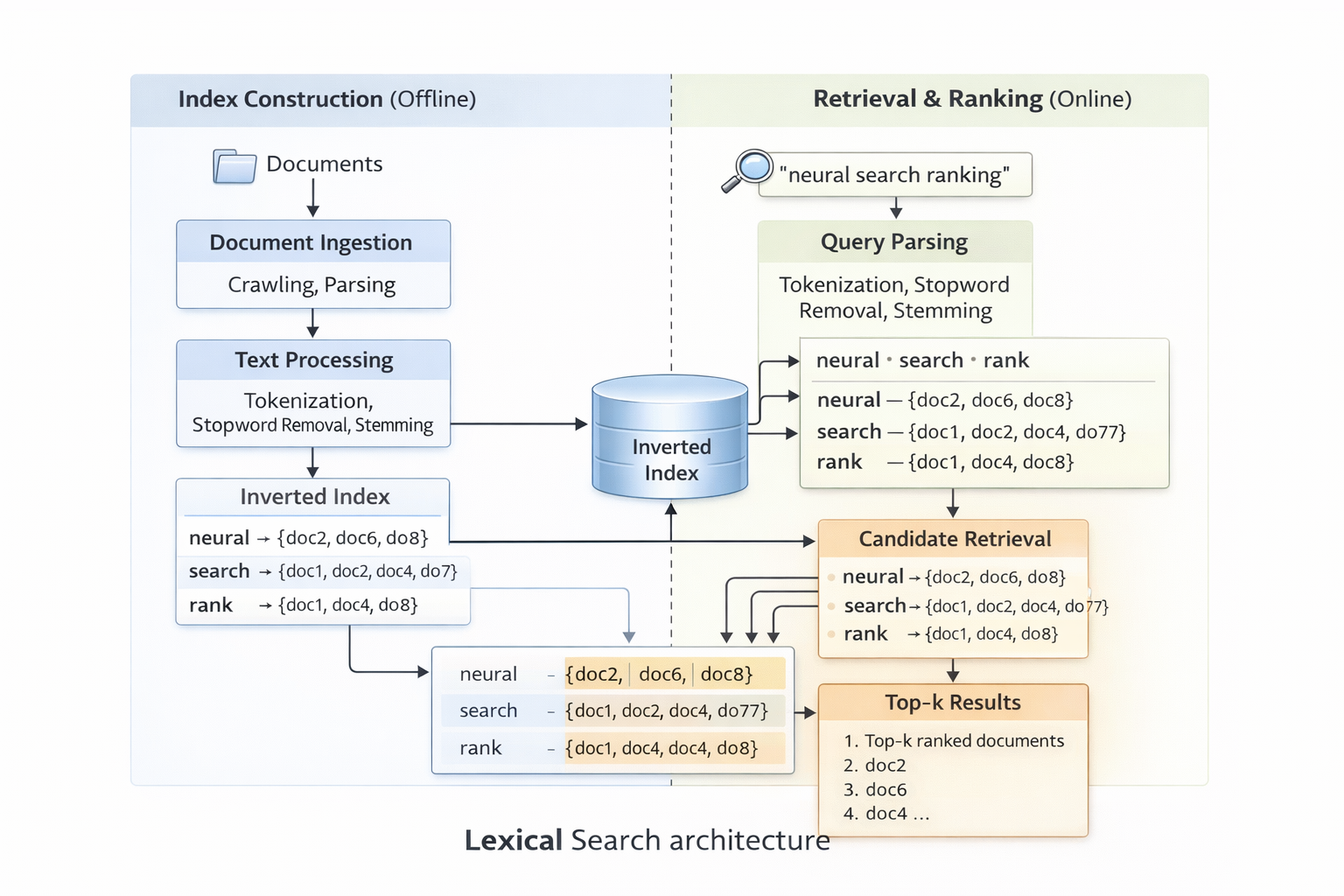}
    \caption{Document Processing for Lexical Search}
    \label{fig:es}
\end{figure*}

These steps normalize text and reduce vocabulary size.

\paragraph{Index Construction}
The processed tokens are stored in an \textit{inverted index}. An inverted index maps each term to the list of documents in which it appears.

Formally, an inverted index can be represented as:

\[
t \rightarrow \{(D_1, f_1), (D_2, f_2), \dots \}
\]

where each entry stores the document identifier and term frequency.

\paragraph{Query Processing}
User queries are processed using the same preprocessing pipeline applied to documents. The system retrieves documents containing the query terms using the inverted index.

\paragraph{Ranking}
The retrieved documents are ranked using scoring functions such as TF-IDF or BM25 to determine the most relevant results.

\subsubsection{Elasticsearch}

Elasticsearch is a distributed search and analytics engine built on top of Apache Lucene. It is designed to efficiently index and search large-scale text data in near real time \cite{gormley2015elasticsearch}. 

\paragraph{Architecture}

Elasticsearch organizes data into several hierarchical components:

\begin{itemize}
\item \textbf{Cluster:} A collection of nodes working together.
\item \textbf{Node:} A single server that stores data and participates in indexing and search operations.
\item \textbf{Index:} A collection of documents with similar characteristics.
\item \textbf{Shard:} A partition of an index that allows data to be distributed across nodes.
\item \textbf{Document:} A JSON object representing a single record.
\end{itemize}

\paragraph{Indexing Process}

When a document is indexed in Elasticsearch, the following steps occur:

\begin{enumerate}
\item The document is received as a JSON object.
\item Text fields are processed by analyzers that perform tokenization and normalization.
\item Tokens are stored in an inverted index.
\item The index is distributed across shards for scalability.
\end{enumerate}

\paragraph{Search Process}

When a user submits a query:

\begin{enumerate}
\item The query is analyzed using the same analyzer used during indexing.
\item Elasticsearch retrieves candidate documents using the inverted index.
\item Each shard computes relevance scores (typically using BM25).
\item Results from all shards are merged and ranked.
\item The top-ranked documents are returned to the user.
\end{enumerate}

\paragraph{Advantages}

Elasticsearch provides several benefits:

\begin{itemize}
\item Near real-time search capabilities
\item Distributed and scalable architecture
\item Full-text search with advanced analyzers
\item Built-in relevance scoring and ranking
\end{itemize}

Due to these properties, Elasticsearch is widely used in applications such as search engines, recommendation systems, log analytics, and enterprise search.

\section{Implementation}
\subsection{Document Parsing Pipeline}

Our document parsing pipeline is designed to efficiently process PDFs while preserving both textual and structural information. Users upload PDF files as raw bytes, and each document is assigned a unique identifier (\texttt{doc\_id}) computed as the MD5 hash of the filename. To ensure stability and performance, the system enforces a file size limit of 100MB.

The processing strategy adapts to document size. PDFs with fewer than 50 pages are processed sequentially in a single thread. For larger documents (50 pages or more), we leverage parallel processing using \texttt{ProcessPoolExecutor}, splitting the document into chunks of approximately 5 pages per worker and using up to \texttt{cpu\_count} workers. This ensures scalable performance for lengthy research papers or technical manuals.

Text extraction is performed on a per-page basis, with two modes. In the default mode, we extract raw text using \texttt{page.get\_text()} from PyMuPDF. When \texttt{use\_llm\_extraction} is enabled, we convert the page to markdown using \texttt{pymupdf4llm.to\_markdown()}, which preserves formatting such as tables, mathematical expressions, and section headings, improving downstream processing for coding agents.

Extracted text is split into fixed-size chunks for indexing and retrieval. Each page is independently divided into segments of 3,000 characters, with a minimum chunk size of 100 characters. Chunks smaller than 100 characters are discarded. The splitting strategy is hard, with no overlap or sentence boundary awareness, ensuring straightforward alignment between chunks and page locations.

Figures and tables are extracted in parallel with text processing. Images smaller than 100×100 pixels are ignored. Captions are detected using spatial heuristics, considering text within 100 pixels of the image bounding box. Tables are identified using PyMuPDF's \texttt{find\_tables()} function and converted into a simple Markdown-style format (\texttt{col1 | col2 | col3}) for indexing and retrieval.

The extracted content is stored across four Elasticsearch indices to support structured queries. The \texttt{documents} index contains metadata such as filename, upload time, total pages, and status. \texttt{document\_chunks} stores per-page text segments, with flags indicating whether text has been processed. \texttt{document\_figures} and \texttt{document\_tables} store figure captions, OCR text, table captions, and table content, enabling multi-modal retrieval.

Search over the document corpus uses a three-tier fallback strategy. Queries first attempt \texttt{match\_phrase} searches for exact phrases, followed by \texttt{match} queries requiring all words (\texttt{AND}), and finally \texttt{match} queries where any word may match (\texttt{OR}). Results are aggregated by \texttt{parent\_doc\_id}, returning matched pages, text snippets (200 characters around the match), figures, and tables, allowing coding agents to access comprehensive context efficiently.

\begin{algorithm}
\caption{Document Parsing Pipeline}
\begin{algorithmic}[1]

\Require PDF file as raw bytes
\Ensure Indexed document data in Elasticsearch

\State Compute \texttt{doc\_id} (MD5 of filename); validate size $\leq$ 100MB
\State Load document and determine total pages $N$

\If{$N < 50$}
    \State Process pages sequentially
\Else
    \State Split into $\sim$5-page chunks and process in parallel using CPU workers
\EndIf

\If{\texttt{use\_llm\_extraction}}
    \State Extract text as Markdown
\Else
    \State Extract raw text per page
\EndIf

\State Split text into 3000-character chunks; discard chunks $<100$ characters
\State Extract figures; detect captions via spatial proximity
\State Detect tables and convert to Markdown format
\State Store metadata, text chunks, figures, and tables in Elasticsearch indices

\end{algorithmic}
\end{algorithm}
\subsection{Tool Calling}

\subsection{Document Search Tool}

As motivated in Section 2, our critical information is not publicly accessible, it creates a gap between the agent’s reasoning capability and the data it can access. Document search tool is motivated by the need to bridge this gap by enabling direct retrieval from domain-specific sources that are otherwise unreachable through conventional search tools.

Search tool is a retrieval abstraction that allows an agent to send queries through a standardized interface. In practice, this mechanism is used when queries require internal or specialized knowledge. For example, when a developer asks, ``Get information on synchronization signal of 5G standards'', the agent can call our document search endpoint:
\begin{verbatim}
POST /search
{ "query": "Find synchronization signals in 38.331" }
\end{verbatim}
The endpoint returns precise, organization-specific information. This avoids irrelevant external results and demonstrates how custom endpoints improve retrieval accuracy in domain-specific tasks.

From an agent perspective, this tool is exposed as a callable function, typically defined in the system prompt or tool schema. For example:
\begin{verbatim}
search_internal(query: string) -> results
\end{verbatim}
The system prompt must instruct the agent when to use this tool, such as: ``Use searchinternal for queries related to internal systems, APIs, or proprietary data. Prefer it over web search when applicable.'' The agent then decides at runtime whether to call this tool based on the query. Modifying the system prompt is critical; without clear guidance, the agent may default to generic search and ignore the custom endpoint, reducing effectiveness.

\subsubsection{LSP Integration}
LSP-based search utilizes the Language Server Protocol to provide semantic access to codebases, enabling queries based on program structure rather than raw text. It leverages compiler-level information such as symbol definitions, references, and type relationships to support precise navigation and analysis.

The need for semantic search arises from the limitations of traditional text-based methods. In large codebases, simple string matching often produces ambiguous or noisy results, as it cannot distinguish between different uses of the same identifier. LSP-based search addresses this by incorporating context and type information, improving accuracy.

For example, when a user asks, “Where is the function findMedian defined?”, LSP-based search can directly locate the function definition. Similarly, a query like “Where is findMedian used?” returns only valid references, excluding unrelated matches. This demonstrates the advantage of structure-aware retrieval.

However, LSP-based systems depend on the availability and quality of language servers. Some languages have incomplete or unreliable implementations, which can limit functionality. Performance issues may occur in large repositories, and misconfiguration can prevent the system from working correctly. These factors can reduce consistency across environments

\subsection{LSP-Based Semantic Code Search}

Traditional code search in development workflows relies heavily on text-based tools such as \texttt{grep} and \texttt{sed}. These tools operate on raw strings and are widely used because they are fast, simple, and universally available across environments. Developers use \texttt{grep} to locate occurrences of identifiers and \texttt{sed} to transform or filter matching lines. However, these tools lack any understanding of program structure, which leads to ambiguous and noisy results. For example, searching for a function name using \texttt{grep} will also return matches inside comments, strings, or unrelated variables. This limitation becomes more severe in large codebases, where filtering relevant results manually becomes time-consuming and error-prone.

The need for LSP-based semantic search arises from these limitations. While text-based tools are effective for quick pattern matching, they cannot distinguish between different semantic roles of the same identifier. LSP-based systems address this by incorporating compiler-level knowledge of the codebase, including symbol definitions, references, and type information. Instead of treating code as plain text, the system interprets it as a structured program. This enables precise queries such as locating only function definitions or valid references, eliminating irrelevant matches. As a result, LSP complements rather than replaces traditional tools by providing accuracy where text-based methods fall short.

LSP-based semantic code search leverages the Language Server Protocol to expose structured code intelligence through a set of queryable operations. The language server maintains an indexed representation of the codebase that is continuously updated as files change. This index includes mappings between symbols and their definitions, as well as relationships such as references and type hierarchies. The agent interacts with this system through operations like \texttt{go\_to\_definition}, \texttt{find\_references}, and \texttt{document\_symbols}. Unlike \texttt{grep}, which returns lines of text, these operations return precise locations and structured metadata. This allows the agent to reason about code in a way that aligns with how compilers and IDEs interpret it.

In practice, LSP-based search is used for tasks that require semantic precision. For example, when a user asks, ``Where is the function \texttt{getUser} defined?'', a \texttt{grep} search would return all textual matches, including irrelevant ones. In contrast, an LSP query such as:
\begin{verbatim}
lsp_definition("getUser")
\end{verbatim}
returns the exact definition location. Similarly, to find usages of the function, the agent can invoke:
\begin{verbatim}
lsp_references("getUser")
\end{verbatim}
which returns only valid references across the codebase. In many workflows, developers may combine both approaches: using \texttt{grep} for broad exploration and LSP for precise navigation. This demonstrates how LSP enhances existing workflows rather than replacing them.

From an agent perspective, LSP capabilities are exposed as callable tools that can be selected based on query intent. These tools are defined in the system prompt or tool schema, for example:
\begin{verbatim}
lsp_definition(symbol: string) -> location
lsp_references(symbol: string) -> list[location]
\end{verbatim}
The system prompt must explicitly guide the agent on when to use LSP versus text-based search, such as: ``Use LSP tools for questions about code structure, definitions, or references. Use text search tools like grep for broad pattern matching or when semantic tools are unavailable.'' This distinction is critical for effective tool use. Without clear instructions, the agent may misuse LSP for simple tasks or rely on text search when precision is required. Proper integration ensures that both approaches are used in a complementary manner, balancing speed and accuracy.

\subsection{Skill Library--Driven Research Workflows}

Modern research-oriented agent systems require more than isolated retrieval capabilities. While document search enables access to external knowledge and LSP-based tools provide precise code understanding, neither alone defines how a task should be solved. The core challenge is not just retrieving information but structuring the process of investigation. For example, answering a systems question may require reading documentation, inspecting code, validating assumptions, and synthesizing results. Without an explicit mechanism to organize these steps, the agent operates in an ad-hoc manner. This motivates the need for a layer where the research process itself can be defined and controlled.

We define the skill library as this orchestration layer, where each skill represents a reusable unit of work, and workflows are constructed by composing these skills. Unlike document search or LSP, which provides specific capabilities, the skill library defines \textit{how} those capabilities are used together.  In this sense, the skill library acts as a programmable interface for defining research procedures.

The key contribution of this layer is that it allows researchers to encode their research methodology directly into the system. For instance, a researcher studying code behavior can define a workflow that first queries documentation, then uses LSP to locate relevant functions, and finally aggregates findings into a structured explanation. Another researcher may define a different workflow that prioritizes empirical validation or cross-referencing multiple sources. These workflows are not hardcoded into the agent but are instead defined as skills, allowing systematic variation and experimentation. As a result, the skill library becomes the primary mechanism for tailoring the agent to different research curricula.

A concrete example better illustrates the role of the skill layer as a research workflow rather than a simple tool chain. Consider a task such as: ``Design and simulate synchronization signals in a communication system.'' This is not a single-step query but a structured research problem that requires iterative information gathering, planning, and implementation.

The workflow begins with a document-centric exploration phase. The agent first performs document search with query expansion, for example by augmenting the initial query with related terms such as modulation schemes, timing recovery, and signal detection. Based on the initial results, the agent iteratively refines its queries, retrieving relevant papers, technical documentation, and reference implementations. This stage is not a one-shot retrieval but a loop where results inform subsequent searches, allowing the agent to build a comprehensive set of references.

Once sufficient material is collected, the workflow transitions to a synthesis phase. The agent aggregates the retrieved information and constructs a structured research plan. This includes identifying the main objective (e.g., evaluating synchronization accuracy), defining subproblems (signal generation, noise modeling, detection), and outlining the abstraction level of the simulation. The agent also determines preliminary design choices such as signal types, parameter ranges, and evaluation metrics. At this stage, the output is not code but a coherent plan describing what needs to be implemented and why.

Following this, the workflow moves to the implementation planning stage. The agent specifies what code needs to be written, including the simulation components (e.g., signal generator, channel model, receiver algorithm). It also determines practical considerations such as logging strategy, computational resources, and reproducibility requirements. For example, the agent may decide to use specific libraries for numerical computation, define how intermediate results are stored, and select visualization tools for analyzing outputs. These decisions ensure that the implementation is aligned with the research objectives.

Finally, the workflow invokes coding tools to implement the planned system. At this stage, the agent generates code based on the previously defined structure, rather than ad-hoc generation. The workflow ensures that the code reflects the research plan, includes appropriate instrumentation (e.g., logging and metrics), and supports experimentation through configurable parameters. This demonstrates that the skill layer is not merely executing isolated tool calls, but orchestrating a full research process from information gathering to implementation.

\subsubsection{Skill Library as a Research Workflow Layer}

A skill is a researcher defined workflow that specifies how the agent should conduct a class of tasks. Unlike primitive tools such as document search or LSP, a skill does not represent a single capability. Instead, it defines an ordered procedure that combines multiple tools, intermediate reasoning steps, and execution constraints. This makes the skill library the layer where researchers can encode methodology rather than only expose functionality.

Researchers write skills as independent workflow specifications. These skills may be stored as prompt templates, structured tool definitions, or external workflow files, depending on the system design. The key idea is that the researcher does not directly hardcode one answer, but instead defines how the agent should approach a problem. In this way, the skill library becomes the place where research curricula, analysis procedures, and coding policies can be tailored.

The agent uses a skill by selecting it for a given task and then executing its steps in sequence. The skill can require the agent to search documents first, inspect code second, construct a plan third, and only then invoke coding tools. This separation is important because it allows the workflow itself to be reused across tasks while the input query changes.

Algorithm~\ref{alg:research_code_flow} defines the role of the skill library at the system level. The important point is that the researcher writes the workflow before runtime, and the agent follows it during runtime. This gives the researcher control over process, not just output.

\paragraph{Example: Code-oriented research workflow.}
A concrete example is a skill for the task: \textit{design and simulate synchronization signals}. In this case, the skill should enforce that the agent does not immediately generate code. Instead, it must first collect references, build a research plan, define simulation conditions, and then move to implementation.

\begin{algorithm}[t]
\caption{Example skill for research-driven code generation}
\label{alg:research_code_flow}
\begin{algorithmic}[1]
\Require research query $q =$ ``design and simulate synchronization signals''
\Ensure research plan, implementation plan, and generated code
\State Expand the initial query into related search queries
\State Run document search using the expanded queries
\State Inspect retrieved results and extract additional keywords
\State Iteratively re-run search until sufficient references are collected
\State Aggregate references and identify main methods, assumptions, and prior designs
\State Write a research plan:
\Statex \hspace{1em} \textbf{Objective:} what problem is being solved
\Statex \hspace{1em} \textbf{Abstraction:} what level of simulation is needed
\Statex \hspace{1em} \textbf{Subtasks:} signal generation, channel model, detection, evaluation
\State Write an implementation plan:
\Statex \hspace{1em} modules to implement
\Statex \hspace{1em} simulation parameters
\Statex \hspace{1em} simulation conditions
\State Define coding policies:
\Statex \hspace{1em} logging strategy
\Statex \hspace{1em} compute resources available
\Statex \hspace{1em} numerical libraries to use
\Statex \hspace{1em} visualization libraries to use
\State Invoke coding tools only after the above steps are complete
\State Generate code for the simulation
\State Return code together with plan, parameters, and execution assumptions
\end{algorithmic}
\end{algorithm}

Algorithm~\ref{alg:research_code_flow} is a concrete instance of the more general skill abstraction. Here, the skill does not merely call tools; it enforces a research procedure. The workflow begins with document search, continues through iterative reference collection and planning, and only then permits implementation. This is the key distinction: the skill library is where researchers define \emph{how} research should be conducted, while the tools execute individual steps within that workflow.
\bibliographystyle{ACM-Reference-Format}
\bibliography{references}
\end{document}